\documentclass[11pt]{article}

\usepackage[final]{acl}

\usepackage{times}
\usepackage{latexsym}

\usepackage[T1]{fontenc}

\usepackage[utf8]{inputenc}

\usepackage{microtype}

\usepackage{inconsolata}

\usepackage{graphicx}
\usepackage[most]{tcolorbox}

\PassOptionsToPackage{table}{xcolor}

\usepackage{makecell}
\usepackage[table]{xcolor}
\usepackage{colortbl}
\usepackage{booktabs}
\usepackage{threeparttable}
\usepackage{multirow}
\usepackage{arydshln}
\usepackage{enumitem}
\setlist[itemize]{leftmargin=*, itemsep=1pt, topsep=2pt}

\definecolor{baseblue}{RGB}{210,230,255}
\definecolor{gptgreen}{RGB}{210,240,210}
\definecolor{synthorange}{RGB}{255,226,185}
\definecolor{humanpurple}{RGB}{226,210,245}
\definecolor{humangray}{RGB}{225,225,225}
\definecolor{sectiongray}{RGB}{205,210,215}

\definecolor{mrshade}{gray}{0.93}

\definecolor{metricblue}{RGB}{72,135,190}
\newcommand{\metriccell}[2]{\cellcolor{metricblue!#1}#2}

\title{When Less Is More: An Empirical Study of Minimal Responses \\ in Counseling Dialogues and the Behavior of LLMs}

\author{Zhiyang Qi \\
  The University of Tokyo \\
  \texttt{zhiyangqi@g.ecc.u-tokyo.ac.jp} }

\begin{document}
\maketitle
\begin{abstract}


In psychological counseling, effective support is not always delivered through long, information-rich responses. Minimal responses, such as backchannel cues and concise empathic statements, help convey attentive listening, express empathy, and encourage clients to continue expressing themselves. However, existing counseling dialogue systems and evaluation frameworks often favor explicit, content-rich replies, overlooking the interactional value of brief counselor utterances. This paper presents a systematic cross-lingual analysis of minimal responses across multiple counseling dialogue datasets. We develop a two-stage filtering method based on utterance length and content, followed by contextual verification using a large language model (LLM). Our analysis shows that minimal responses are common in human-collected datasets but substantially underrepresented in LLM-generated ones. We further evaluate current LLMs in manually curated dialogue contexts where human counselors used minimal responses. The results show that strong commercial LLMs are capable of generating minimal responses when explicitly instructed, but still struggle to determine when such responses are appropriate. Counseling-specific models trained on synthetic data perform particularly poorly, tending instead to produce longer and more information-rich responses. Moreover, LLM-based response-quality evaluation may undervalue minimal responses, even when they are interactionally appropriate.

\end{abstract}

\section{Introduction}

\begin{figure}[t!]
  \centering
  \includegraphics[width=\linewidth]{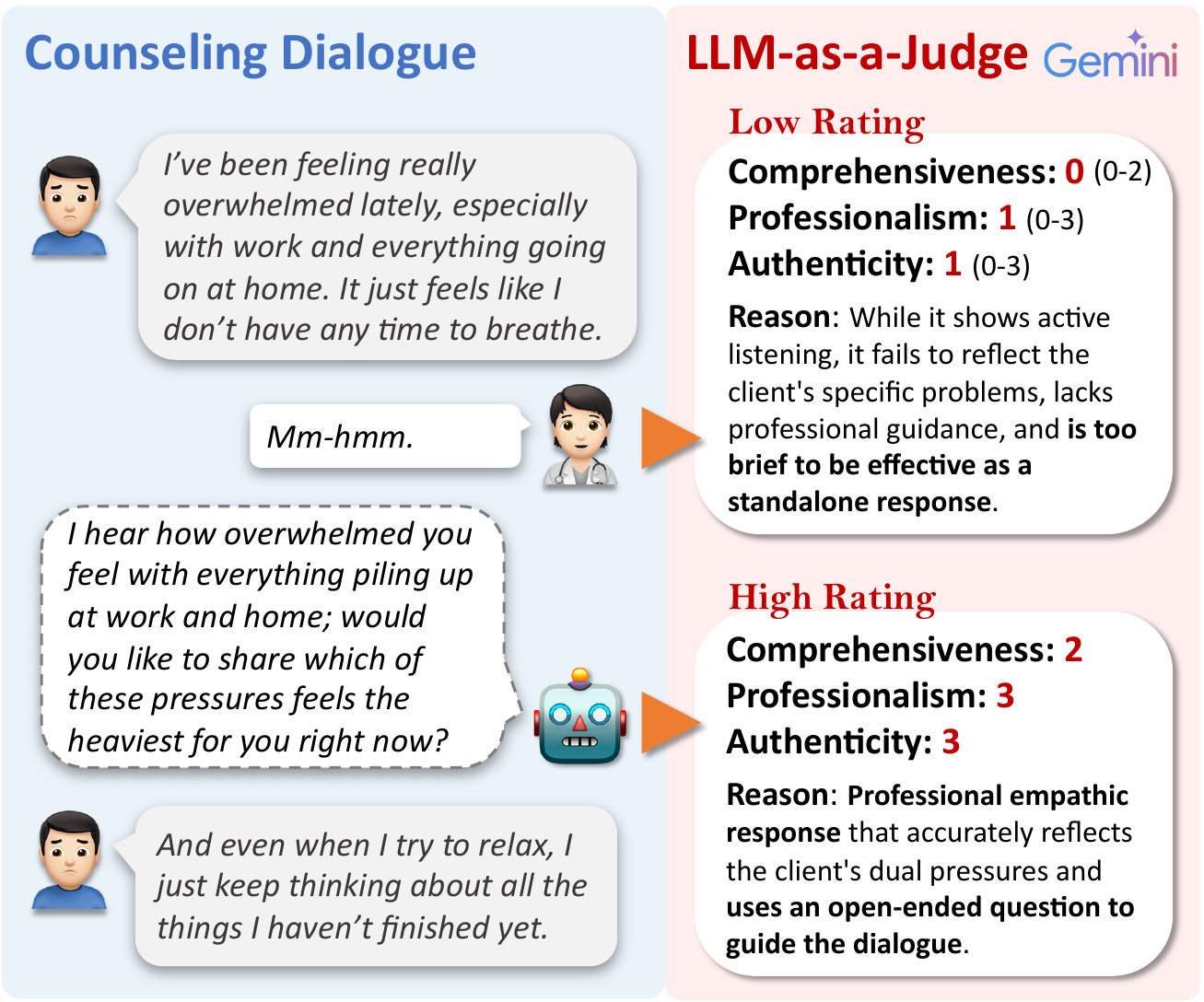}
  \caption{Minimal responses are a key technique in real counseling. However, current LLM-based evaluation methods (here using \citet{zhang-etal-2024-cpsycoun}’s prompt with Gemini) tend to undervalue such responses due to their brevity, favoring more elaborate replies.}
  \label{fig:fig_intro}
\end{figure}

Mental health issues have attracted increasing attention in recent years \cite{world2022mental, raising}, yet many individuals still lack timely access to professional support due to the shortage of counselors and the high cost of services. This has motivated growing interest in large language models (LLMs) as virtual counselors \cite{liu2023chatcounselor, qi2025emostage, chen-etal-2023-soulchat}, making high-quality counseling dialogue data essential for both training and evaluation.

Collecting real counseling dialogues, however, is challenging because they contain sensitive and private information. Existing public datasets therefore rely on indirect collection strategies, such as transcribing public counseling videos (AnnoMI \cite{AnnoMi}), collecting role-play dialogues with trained counselors (KokoroChat \cite{qi-etal-2025-kokorochat}), or anonymizing online counseling conversations by rewriting client utterances (PsyDial \cite{qiu-lan-2025-psydial}). Since such human-collected resources remain limited in scale, recent work has increasingly turned to LLM-synthesized counseling datasets, including SoulChat \cite{chen-etal-2023-soulchat}, SmileChat \cite{qiu-etal-2024-smile}, and C\textsc{actus} \cite{lee-etal-2024-cactus}. In parallel, several evaluation frameworks have been proposed for counseling response generation, typically assessing dimensions such as informativeness, professionalism, realism, and safety \cite{zhang-etal-2024-cpsycoun, zhao-etal-2024-esc}.

While these datasets and evaluation methods have advanced counseling dialogue research, they often emphasize informative and structurally complete responses. This focus risks overlooking \textbf{minimal responses}, brief but interactionally important counselor utterances that play a central role in counseling practice. Examples such as ``uh-huh'' and ``I see'' are considered essential micro-skills: although they convey little propositional content, they express empathy, acceptance, and attentiveness without interrupting the client's narrative, thereby encouraging further self-disclosure \cite{ivey2018intentional, hill2019helping}. Rather than providing information or direction, minimal responses function as continuers \cite{sacks1974simplest, schegloff1982discourse}, facilitating client expression and supporting the therapeutic alliance in client-centered counseling \cite{rogers1957necessary}.

In contrast, we observe that counselor utterances in some LLM-generated counseling datasets often follow fixed, information-dense patterns, such as expressing empathy followed by a question\footnote{For example, according to our analysis, 43.9\% of the 24,220 counselor utterances across 3,084 dialogues in CPsyCounD \cite{zhang-etal-2024-cpsycoun} end with a question mark. In PsyDTCorpus \cite{xie-etal-2025-psydt}, the proportion is 67.4\% among 86,054 counselor utterances across 4,760 dialogues.}, as illustrated in Figure~\ref{fig:fig_intro}.
Although questions can help advance counseling conversations, their immediate effects are mixed, and frequent questioning may be associated with lower perceived empathy, helpfulness, and session smoothness \citep{williams2023questions}.
Such response patterns may be inherited by models fine-tuned on synthetic data, leading them to produce verbose or templated responses.
Moreover, utterance-level LLM-based evaluation may further reinforce this tendency by favoring informative and well-structured responses over brief but interactionally meaningful minimal responses.

Motivated by the above observations, we investigate minimal responses in psychological counseling dialogues and make two contributions:

\begin{itemize}
    \item \textbf{Cross-lingual and cross-dataset analysis.}
    We systematically analyze minimal responses across seven human-collected and LLM-generated counseling dialogue datasets in Chinese, Japanese, and English. To support this analysis, we develop a two-stage filtering pipeline that combines rule-based screening based on utterance length and content with LLM-based contextual verification. The results show that minimal responses are substantially more common in human-collected datasets than in LLM-generated ones.

    \item \textbf{Multi-model evaluation in minimal-response contexts.}
    Using manually curated dialogue contexts in which human counselors used minimal responses, we systematically evaluate multiple models across languages. We find that
    (1) models trained on human counseling data are more likely to generate minimal responses than those trained on synthetic data;
    (2) even strong commercial LLMs can generate minimal responses when explicitly instructed, but still struggle to determine when they are appropriate; and
    (3) LLM-based response-quality evaluation may favor more content-rich responses and undervalue interactionally appropriate minimal responses.
\end{itemize}

These findings highlight the need to consider minimal responses in counseling data construction, model development, and evaluation.

\section{Dataset Analysis}


\begin{table*}[t!]
\centering
\footnotesize
\setlength{\tabcolsep}{2.8pt}
\renewcommand{\arraystretch}{1.18}

\resizebox{\textwidth}{!}{
\begin{tabular}{
    l c r c r r r
    >{\columncolor{mrshade}}r
    >{\columncolor{mrshade}}r
    r
}
\toprule

\multicolumn{6}{c}{\textbf{Dataset Statistics}}
& \multicolumn{1}{c}{\textbf{(1) Rule-based Filtering}}
& \multicolumn{3}{c}{\textbf{(2) LLM-based Contextual Verification}} \\

\cmidrule(lr){1-6}
\cmidrule(lr){7-7}
\cmidrule(lr){8-10}

\multirow{2}{*}{\textbf{Dataset}}
& \multirow{2}{*}{\textbf{Lang.}}
& \multirow{2}{*}{\textbf{\#Dial.}}
& \multirow{2}{*}{\textbf{\#Couns.}}
& \multirow{2}{*}{
    \makecell{
        \textbf{Avg. Utter.}\\
        \textbf{per Dialogue}
    }
}
& \multirow{2}{*}{
    \makecell{
        \textbf{\#Counselor}\\
        \textbf{Utterances}
    }
}
& \multirow{2}{*}{
    \makecell{
        \textbf{Rule-filtered}\\
        \textbf{Candidates}\\
        \textbf{(\# / \%)}
    }
}
& \multicolumn{2}{c}{
    \cellcolor{mrshade}\textbf{Minimal Responses}
}
& \multirow{2}{*}{
    \makecell{
        \textbf{Other Short}\\
        \textbf{Responses}\\
        \textbf{(\# / \%)}
    }
} \\

\cmidrule(lr){8-9}

&
&
&
&
&
&
&
\makecell{
    \textbf{Backchannel-like}\\
    \textbf{Responses}\\
    \textbf{(\# / \%)}
}
&
\makecell{
    \textbf{Brief Empathic/}\\
    \textbf{Reflective Responses}\\
    \textbf{(\# / \%)}
}
& \\

\midrule

\multicolumn{10}{l}{\textit{LLM-generated datasets}} \\

C\textsc{actus} \cite{lee-etal-2024-cactus}
& EN
& 31,577
& --
& 30.53
& 491,304
& 4 {\small (0.00\%)}
& 0 {\small (0.00\%)}
& 3 {\small (0.00\%)}
& 1 {\small (0.00\%)} \\

CPsyCounD \cite{zhang-etal-2024-cpsycoun}
& ZH
& 3,084
& --
& 15.94
& 24,220
& 37 {\small (0.15\%)}
& 0 {\small (0.00\%)}
& 1 {\small (0.00\%)}
& 36 {\small (0.15\%)} \\

PsyDTCorpus \cite{xie-etal-2025-psydt}
& ZH
& 4,760
& --
& 36.16
& 86,054
& 1 {\small (0.00\%)}
& 0 {\small (0.00\%)}
& 1 {\small (0.00\%)}
& 0 {\small (0.00\%)} \\

SmileChat \cite{qiu-etal-2024-smile}
& ZH
& 55,165
& --
& 11.38
& 309,698
& 183 {\small (0.06\%)}
& 1 {\small (0.00\%)}
& 68 {\small (0.02\%)}
& 114 {\small (0.04\%)} \\

\addlinespace[3pt]
\multicolumn{10}{l}{\textit{Human-collected datasets}} \\

AnnoMI \cite{AnnoMi}
& EN
& 133
& N/R
& 72.64
& 4,859
& 1,184 {\small (24.37\%)}
& \textbf{877 {\small (18.05\%)}}
& \textbf{128 {\small (2.63\%)}}
& 179 {\small (3.68\%)} \\

PsyDial-D4 \cite{qiu-lan-2025-psydial}
& ZH
& 2,382
& 40
& 75.59
& 90,031
& 4,741 {\small (5.27\%)}
& \textbf{861 {\small (0.96\%)}}
& \textbf{2,778 {\small (3.09\%)}}
& 1,102 {\small (1.22\%)} \\

KokoroChat \cite{qi-etal-2025-kokorochat}
& JA
& 6,589
& 424
& 71.63
& 237,735
& 12,161 {\small (5.12\%)}
& \textbf{7,555 {\small (3.18\%)}}
& \textbf{3,747 {\small (1.58\%)}}
& 859 {\small (0.36\%)} \\

\bottomrule
\end{tabular}
}

\caption{
Dataset statistics and short-response distributions obtained using the
two-stage identification pipeline.
In Step~(1), candidates are identified through rule-based filtering based on
utterance length and content.
In Step~(2), GPT-5.4-mini verifies each candidate using its dialogue context and classifies it into one of three categories: backchannel-like responses, brief empathic/reflective responses, or other short responses.
In this study, the first two categories are collectively considered minimal
responses and are highlighted in gray.
Percentages are calculated relative to the total number of counselor
utterances.
``N/R'' indicates that the number of counselors was not reported, whereas
``--'' indicates that the number is not applicable to LLM-generated datasets.
}
\label{tab:dataset_stats}

\end{table*}

Since minimal responses in existing counseling dialogue datasets have not been
systematically examined, this section analyzes seven datasets spanning Chinese,
Japanese, and English, as summarized in
Table~\ref{tab:dataset_stats}.\footnote{For datasets containing consecutive utterances from the same speaker,
adjacent utterances were merged before calculating statistics and conducting
the analysis.} They include four LLM-generated datasets,
CACTUS~\cite{lee-etal-2024-cactus},
CPsyCounD~\cite{zhang-etal-2024-cpsycoun},
PsyDTCorpus~\cite{xie-etal-2025-psydt}, and
SmileChat~\cite{qiu-etal-2024-smile}, and three human-collected datasets,
PsyDial-D4~\cite{qiu-lan-2025-psydial},
AnnoMI~\cite{AnnoMi}, and
KokoroChat~\cite{qi-etal-2025-kokorochat}.
Although conventional backchannels such as ``uh-huh'' and ``I see'' are prototypical minimal responses, brief empathic or reflective utterances such as ``that sounds difficult'' can serve a similar role by acknowledging the client's experience without introducing a new question or topic. We therefore include both types in our operational definition of minimal responses.

To identify such responses more comprehensively, we adopt a two-stage procedure that combines rule-based filtering with LLM-based contextual verification. First, we apply rule-based filtering to counselor utterances using length thresholds and predefined keyword lists. Because minimal responses often encourage clients to continue speaking and tend to occur within longer client narratives, we also require the immediately following client utterance to exceed a predefined length threshold. Specifically, counselor utterances are limited to 15 characters in Chinese and Japanese and 10 words in English, while the following client utterance must contain at least 10 characters in Chinese and Japanese or 5 words in English. These relatively permissive thresholds are intended to reduce false negatives. We then apply a content-based filter using predefined keyword lists to remove short questions, conversation-advancing prompts, directives or suggestions, and utterances such as thanks, apologies, greetings, polite closings, and simple encouragements, whose primary function is not to facilitate further client expression. The keyword lists are shown in Figure~\ref{fig:rule_filter_patterns}, and the full filtering criteria are provided in Appendix~\ref{app:rule_based_filtering}.

Second, to improve precision, we use GPT-5.4-mini\footnote{\url{https://developers.openai.com/api/docs/models/gpt-5.4-mini}} to verify each remaining candidate based on its dialogue context. The model classifies each candidate into one of three categories: (1) backchannel-like responses, (2) brief empathic or reflective responses, and (3) other short responses. The classification prompt is shown in Figure~\ref{fig:contextual_classification_prompt} in Appendix~\ref{app:rule_based_filtering}. We further manually inspected 100 randomly sampled instances from each category in each of the three human-collected datasets, and all category-level validation rates exceeded 94\%, supporting the reliability of the classification procedure.

Table~\ref{tab:dataset_stats} presents the results of this two-stage procedure. Minimal responses are almost absent from the LLM-generated datasets but occur substantially more often in all three human-collected datasets. AnnoMI shows the highest proportion, possibly because it is transcribed from face-to-face counseling, where interactive feedback is more frequent. By contrast, PsyDial-D4 and KokoroChat consist of text-based online counseling dialogues and contain lower proportions of minimal responses.



\section{Experiment}

\subsection{Experimental Setup}
To examine whether current models can use minimal responses appropriately, we conduct response-generation experiments across multiple languages. We manually selected a subset of dialogue contexts in which counselors used minimal responses from the human-collected Chinese, Japanese, and English datasets. We then validated these examples using both Gemini-3.1-Flash-Lite\footnote{\url{https://ai.google.dev/gemini-api/docs/models/gemini-3.1-flash-lite}} and GPT-5.4-mini to ensure that they were appropriate for the evaluation. After validation, we retained 284, 300, and 299 dialogue history--response pairs for Chinese, Japanese, and English, respectively. Figure~\ref{fig:minimal_response_examples} in Appendix~\ref{app:experimental_details} presents representative examples. We provide each dialogue history to the evaluated models and ask them to generate the next counselor response.

For models not specifically developed for psychological counseling, we use two prompting settings:
\begin{itemize}
    \item \textbf{General prompt:} The model is asked to generate the next counselor response based on the dialogue history. This setting evaluates whether the model naturally produces a minimal response from context, thereby testing its ability to determine when such a response is appropriate.
    
    \item \textbf{Instructional prompt:} The model is explicitly instructed to prioritize a minimal response when the client is still in the process of expressing emotions or experiences. This setting evaluates whether the model can follow an explicit instruction and generate a minimal response, thereby testing its ability to produce such responses.
\end{itemize}

We compare general-purpose open-source models such as Qwen \cite{qwen} and Llama \cite{llama3}; GPT-5.4\footnote{\url{https://developers.openai.com/api/docs/models/gpt-5.4}}, a state-of-the-art commercial model as of late April 2026; and counseling-specific models including SoulChat2.0 \cite{xie-etal-2025-psydt}, MindChat\footnote{\url{https://huggingface.co/X-D-Lab/MindChat-Qwen-7B-v2}}, KokoroChat-Full \cite{qi-etal-2025-kokorochat}, and PsyDial-Pi4 \cite{qiu-lan-2025-psydial}\footnote{PsyDial-Pi4 and KokoroChat-Full were retrained following the settings of their original papers after removing all complete conversations containing any test case.}. We also include the original human counselor responses as a reference. Details of the models and prompts are provided in Appendix~\ref{app:models_prompts}.

\begin{table*}[t]
\centering
\scriptsize
\setlength{\tabcolsep}{3.5pt}
\renewcommand{\arraystretch}{1.18}

\begin{threeparttable}
\begin{tabular}{
    >{\raggedright\arraybackslash}p{0.34\textwidth}
    c c c c c c c c
}
\toprule

\multirow{2}{*}{\textbf{Model}}
& \multirow{2}{*}{\makecell{\textbf{Avg.}\\\textbf{Length}}}
& \multicolumn{5}{c}{\textbf{Response-quality Evaluation}}
& \multicolumn{2}{c}{\textbf{Minimal-response Evaluation}} \\

\cmidrule(lr){3-7}
\cmidrule(lr){8-9}

&
& \textbf{Comp.}
& \textbf{Prof.}
& \textbf{Auth.}
& \textbf{Safe.}
& \textbf{Quality Avg.}
& \textbf{MR (\%)}
& \textbf{Interrupt. $\downarrow$} \\

\midrule

\rowcolor{sectiongray}
\multicolumn{9}{l}{
    \textit{Chinese dataset: PsyDial-D4 ($N=284$)}
} \\

\rowcolor{baseblue}
Qwen3-8B (general prompt)
& 76.15
& 1.69 & 2.21 & 2.33 & 0.99
& \metriccell{56}{1.81}
& \metriccell{8}{0.00}
& \metriccell{14}{3.54} \\

\rowcolor{baseblue}
Qwen3-8B (instructional prompt)
& 24.24
& 0.77 & 1.04 & 1.52 & 0.96
& \metriccell{30}{1.07}
& \metriccell{23}{23.59}
& \metriccell{41}{1.83} \\

\rowcolor{gptgreen}
GPT-5.4 (general prompt)
& 132.05
& 2.00 & 2.96 & 2.97 & 1.00
& \metriccell{70}{2.23}
& \metriccell{8}{0.00}
& \metriccell{13}{3.62} \\

\rowcolor{gptgreen}
GPT-5.4 (instructional prompt)
& 8.52
& 0.06 & 0.12 & 0.50 & 0.97
& \metriccell{8}{0.41}
& \metriccell{68}{97.18}
& \metriccell{69}{0.10} \\

\rowcolor{synthorange}
SoulChat2.0
& 52.10
& 1.21 & 1.58 & 1.88 & 0.99
& \metriccell{42}{1.42}
& \metriccell{8}{0.00}
& \metriccell{25}{2.89} \\

\rowcolor{synthorange}
MindChat
& 58.11
& 0.97 & 0.97 & 1.27 & 0.94
& \metriccell{29}{1.04}
& \metriccell{8}{0.35}
& \metriccell{16}{3.43} \\

\rowcolor{humanpurple}
PsyDial-Pi4 (retrained)
& 46.81
& 0.94 & 1.25 & 1.67 & 0.99
& \metriccell{36}{1.22}
& \metriccell{16}{12.68}
& \metriccell{41}{1.85} \\

\rowcolor{humangray}
Human
& 5.30
& 0.10 & 0.16 & 0.59 & 0.98
& \metriccell{10}{0.46}
& \metriccell{70}{100.00}
& \metriccell{68}{0.12} \\

\addlinespace[3pt]
\rowcolor{sectiongray}
\multicolumn{9}{l}{
    \textit{Japanese dataset: KokoroChat ($N=300$)}
} \\

\rowcolor{baseblue}
Llama-3.1-Swallow (general prompt)
& 41.29
& 1.20 & 1.62 & 1.82 & 0.97
& \metriccell{42}{1.40}
& \metriccell{10}{2.67}
& \metriccell{29}{2.63} \\

\rowcolor{baseblue}
Llama-3.1-Swallow (instructional prompt)
& 20.32
& 0.71 & 1.13 & 1.47 & 0.95
& \metriccell{30}{1.07}
& \metriccell{30}{34.83}
& \metriccell{46}{1.54} \\

\rowcolor{gptgreen}
GPT-5.4 (general prompt)
& 66.41
& 1.93 & 2.75 & 2.82 & 0.99
& \metriccell{66}{2.12}
& \metriccell{8}{0.00}
& \metriccell{26}{2.81} \\

\rowcolor{gptgreen}
GPT-5.4 (instructional prompt)
& 10.05
& 0.23 & 0.52 & 0.86 & 0.91
& \metriccell{15}{0.63}
& \metriccell{59}{82.00}
& \metriccell{66}{0.26} \\

\rowcolor{humanpurple}
KokoroChat-Full (retrained)
& 12.99
& 0.28 & 0.62 & 0.94 & 0.90
& \metriccell{18}{0.69}
& \metriccell{59}{82.67}
& \metriccell{65}{0.32} \\

\rowcolor{humangray}
Human
& 3.80
& 0.10 & 0.40 & 0.70 & 0.89
& \metriccell{12}{0.52}
& \metriccell{70}{100.00}
& \metriccell{70}{0.01} \\

\addlinespace[3pt]
\rowcolor{sectiongray}
\multicolumn{9}{l}{
    \textit{English dataset: AnnoMI ($N=299$)}
} \\

\rowcolor{baseblue}
Llama-3 (general prompt)
& 49.01
& 1.88 & 2.54 & 2.52 & 0.99
& \metriccell{61}{1.98}
& \metriccell{8}{0.00}
& \metriccell{8}{3.94} \\

\rowcolor{baseblue}
Llama-3 (instructional prompt)
& 2.13
& 0.21 & 0.56 & 1.19 & 0.98
& \metriccell{19}{0.73}
& \metriccell{67}{94.98}
& \metriccell{67}{0.18} \\

\rowcolor{humangray}
Human
& 1.08
& 0.17 & 0.49 & 1.09 & 0.98
& \metriccell{17}{0.68}
& \metriccell{70}{100.00}
& \metriccell{70}{0.02} \\

\bottomrule
\end{tabular}

\caption{
Results of minimal-response generation experiments across three languages.
Response quality is evaluated by comprehensiveness (Comp.),
professionalism (Prof.), authenticity (Auth.), and safety (Safe.),
with Quality Avg.\ denoting their average.
MR is the percentage of minimal responses, while Interrupt.\ measures
the likelihood of interrupting the client's continued expression
on a 0--5 scale, where lower is better.
LLM-based scores are averaged over GPT-5.4-mini and Gemini-3.1-Flash-Lite.
Darker metric cells indicate better performance and are normalized
separately for each metric.
Row colors denote model categories:
blue for general-purpose open-source models,
green for GPT-5.4,
orange and purple for models fine-tuned on synthetic and
human-collected counseling data, respectively,
and gray for human responses.
}
\label{tab:minimal_response_generation}

\end{threeparttable}
\end{table*}

\subsection{Evaluation}
To improve evaluation robustness and reduce reliance on a single model, we use both GPT-5.4-mini and Gemini-3.1-Flash-Lite as evaluators. Each generated response is assessed from three perspectives: whether it constitutes a minimal response, its likelihood of interrupting the client's continued expression on a 0--5 scale (lower is better), and its response quality. For response-quality evaluation, we adopt the prompt from prior work~\cite{zhang-etal-2024-cpsycoun}, which assesses comprehensiveness on a 0--2 scale, professionalism and authenticity on 0--3 scales, and safety on a 0--1 scale. The overall quality score is calculated as the average of these four dimensions. By comparing interruption risk with response-quality scores, we examine whether LLM-based evaluation favors content-rich responses even when minimal responses are more appropriate. The evaluation prompts are provided in Appendix~\ref{app:llm_eval_prompt}.

\subsection{Results}

Table~\ref{tab:minimal_response_generation} presents the results of
minimal-response generation and response-quality evaluation across three
languages. We identify three main findings.

\textbf{Models trained on human counseling data generate minimal responses more often than those trained on synthetic data.}
In Chinese, SoulChat2.0 and MindChat, both trained on synthetic counseling
dialogues, achieve MR rates of only 0.00\% and 0.35\%, respectively, whereas
the retrained PsyDial-Pi4 model reaches 12.68\%. In Japanese, the retrained
KokoroChat-Full model achieves 82.67\%. This pattern is consistent with our
earlier observation that minimal responses are nearly absent from synthetic
counseling datasets but substantially more common in human-collected data,
leaving synthetic-data-trained models with few opportunities to learn this
behavior.

\textbf{General-purpose LLMs can generate minimal responses when explicitly
instructed, but do not reliably select them under standard prompting.}
Under general prompting, GPT-5.4 produces no minimal responses in either
Chinese or Japanese, although all test contexts were manually identified as
appropriate for one. Explicit instruction raises its MR rate to 97.18\% in
Chinese and 82.00\% in Japanese. Qwen3-8B similarly increases from 0.00\% to
23.59\%, and Llama-3.1-Swallow from 2.67\% to 34.83\%. In English, the
comparatively smaller Llama-3 reaches 94.98\% under the instructional prompt.
This may be due to the spoken-transcript nature of AnnoMI, where certain cues, such as trailing hyphens, clearly indicate that the client has not finished speaking, making it easier for the model to recognize when a minimal response is appropriate.

\textbf{Existing LLM-based response-quality evaluation tends to favor
content-rich responses and may undervalue appropriate minimal responses.}
For GPT-5.4, switching from the general to the instructional prompt reduces
Quality Avg.\ from 2.23 to 0.41 in Chinese and from 2.12 to 0.63 in Japanese,
while improving interruption scores from 3.62 to 0.10 and from 2.81 to 0.26,
respectively. The same pattern appears in English, where Llama-3's Quality
Avg.\ decreases from 1.98 to 0.73 as its interruption score improves from 3.94
to 0.18. Human responses also receive relatively low quality scores of 0.46,
0.52, and 0.68 despite near-zero interruption scores.
To make this relationship more explicit,
Appendix~\ref{app:quality_correlation} presents scatterplots relating
LLM-based quality scores to interruption scores and minimal-response rates.
These results indicate that response-quality criteria reward
comprehensiveness and informativeness but insufficiently capture the
interactional value of brief responses that encourage clients to continue
speaking.

\section{Related Work}

Backchannels have been widely studied in conversation analysis and spoken dialogue systems that signal attention, understanding, or agreement without taking the conversational floor \cite{yngve1970getting,sacks1974simplest,schegloff1982discourse}. Prior work has explored automatic backchannel prediction and generation, including when to produce a backchannel and which form to use \cite{ward2000prosodic,gravano2011turn,kawahara2016prediction,inoue2025yeah}, highlighting their role in making human-human communication smoother and more natural.

Minimal responses in counseling are related to backchannels, but serve more
specific therapeutic functions. Brief acknowledgments and encouragers help
counselors express empathy, acceptance, and attentiveness while allowing clients
to continue exploring their experiences
\cite{ivey2018intentional,hill2019helping}, which is consistent with
client-centered counseling \cite{rogers1957necessary}. Related concepts have
appeared as \textit{Grounding} and \textit{Minimal Encouragement} in prior work \cite{shah2022modeling,li-etal-2023-understanding},
but only as one of many dialogue-act or annotation labels rather than as a
primary focus of study. Consequently, little is
known about how minimal responses are represented in counseling dialogue
datasets or whether LLM-based counseling systems can generate them
appropriately. This work addresses this gap by analyzing their distribution
and evaluating their generation across models and prompting strategies.

\section{Conclusion}

This paper examined minimal responses in psychological counseling dialogues.
We found that they are common in human-collected data but rare in synthetic
datasets, and that models trained on synthetic data struggle to generate them.
Although general-purpose LLMs can produce minimal responses when explicitly
instructed, they do not reliably determine when such responses are appropriate.
Conventional response-quality evaluation also tends to undervalue minimal
responses. These findings highlight the need to better incorporate minimal
responses into counseling data construction, model training, and evaluation.

\section*{Acknowledgments}

We sincerely thank Michimasa Inaba, Associate Professor at The University of Electro-Communications, for providing GPU resources. We also thank the anonymous reviewers for their constructive feedback. This work was supported by JST ERATO (JPMJER2502) and the MEXT Supporting Pioneering Research through AI for 1,000 Discovery Challenges Program (SPReAD), Japan, under Grant Number JPMXP1726302875.

\section*{Limitations}

This study has several limitations. First, cross-dataset comparisons may be
affected by differences in collection modality and counselor characteristics.
AnnoMI consists of transcripts of face-to-face counseling, whereas PsyDial-D4
and KokoroChat are text-based online counseling datasets. These differences may
influence conversational rhythm and the frequency of minimal responses.
Variation in counselors' interaction styles and therapeutic orientations may
also contribute to the observed differences across datasets.

Second, although using two LLM evaluators improves robustness, LLM-based
evaluation remains a preliminary and scalable proxy rather than a substitute
for expert or client evaluation. Human counselor responses provide a useful
reference, but they do not establish the therapeutic appropriateness or
client-perceived helpfulness of the generated responses. Future work should
compare LLM judgments with evaluations from counseling experts and clients.

Finally, our experiments examine minimal responses only in selected local
contexts. We do not test whether incorporating them into a complete, dynamic
counseling process improves the interaction, or what timing and frequency are
most effective. Excessive or repetitive use, such as repeatedly responding
with ``Hmm,'' may feel mechanical or irritating rather than supportive.
Future work should evaluate when and how often minimal responses should be used
over full counseling sessions, ideally with feedback from counseling experts
and clients.

\section*{Ethical Considerations}

This study uses existing publicly available counseling dialogue datasets in accordance with the terms of their respective licenses. To the best of our knowledge, these datasets had undergone anonymization or privacy screening before public release. We do not collect new data from real clients or deploy any system for actual counseling. Since counseling dialogues may contain sensitive information, we analyze the data only at the aggregate level and do not attempt to identify or interpret individual clients.

Our findings should not be interpreted as suggesting that LLMs can provide professional counseling. Minimal responses are only one type of counseling-related interactional behavior, and their appropriateness depends on the broader therapeutic context. This work aims to highlight an overlooked aspect of counseling dialogue modeling and to support the development of safer and more realistic counseling dialogue systems.


\bibliography{custom}

\appendix

\begin{figure*}[t!]
\centering
\includegraphics[width=\textwidth]{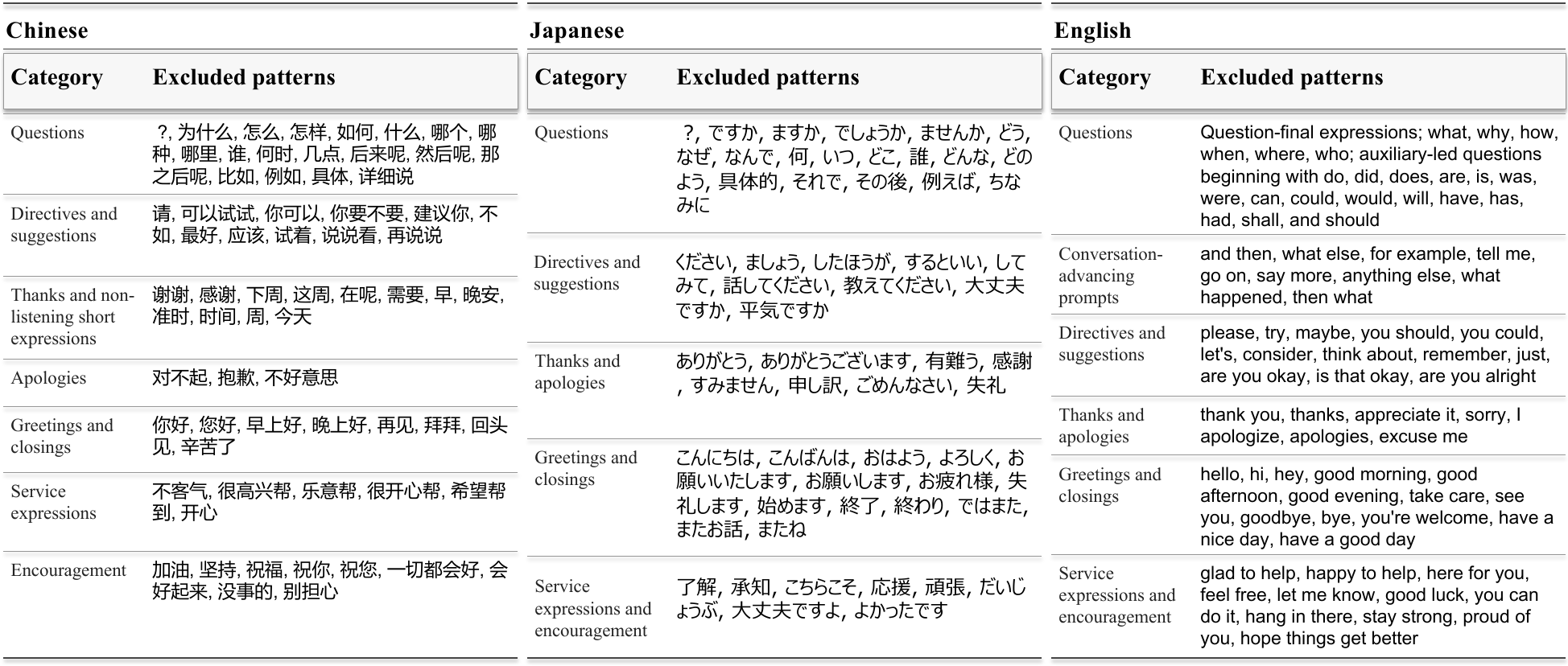}
\caption{Language-specific exclusion patterns used in content-based filtering. The three panels show the original exclusion lists for Chinese, Japanese, and English, respectively.}
\label{fig:rule_filter_patterns}
\end{figure*}

\section{Details of Minimal-Response Identification}
\label{app:rule_based_filtering}

Before identification, adjacent utterances produced by the same speaker are
merged into a single turn. Minimal responses are then identified in two stages:
rule-based candidate filtering followed by LLM-based contextual classification.

In the first stage, rule-based filtering combines language-specific length
thresholds with keyword- and pattern-based exclusion rules. We first retain
short counselor utterances that satisfy the length constraints shown in
Table~\ref{tab:rule_filter_thresholds}. We also require the immediately
following client utterance to exceed a language-specific minimum length. This
requirement serves as a heuristic indicator that the client continued
elaborating after the brief counselor response.

\begin{table}[h!]
\centering
\small
\setlength{\tabcolsep}{4pt}
\renewcommand{\arraystretch}{1.15}
\begin{tabular}{lcc}
\toprule
\textbf{Language}
& \makecell{\textbf{Counselor}\\\textbf{utterance}}
& \makecell{\textbf{Next client}\\\textbf{utterance}} \\
\midrule
Chinese  & $\leq 15$ characters & $\geq 10$ characters \\
Japanese & $\leq 15$ characters & $\geq 10$ characters \\
English  & $\leq 10$ words      & $\geq 5$ words \\
\bottomrule
\end{tabular}
\caption{
Language-specific length thresholds used in rule-based candidate filtering.
The counselor-utterance threshold specifies the maximum length of a candidate
response, while the next-client-utterance threshold specifies the minimum
length of the immediately following client turn.
}
\label{tab:rule_filter_thresholds}
\end{table}

We then apply language-specific keyword and pattern rules to exclude short
utterances whose primary functions clearly differ from those of minimal
responses. As shown in Figure~\ref{fig:rule_filter_patterns}, these rules cover
questions, conversation-advancing prompts, directives, suggestions, thanks,
apologies, greetings, closings, service or logistical expressions, and other
formulaic utterances. Although such responses may be brief, they do not
primarily acknowledge the client's ongoing narrative or leave the
conversational floor open for continued expression.

In the second stage, GPT-5.4-mini contextually classifies each candidate
retained after rule-based filtering. The model jointly considers the recent
dialogue history, the candidate counselor utterance, and the immediately
following client utterance. As shown in
Figure~\ref{fig:contextual_classification_prompt}, each candidate is classified into one of three categories: (1) backchannel-like responses, (2) brief empathic or reflective responses, and (3) other short responses. The first category captures brief
acknowledgments that mainly signal listening and allow the client to continue
speaking. The second captures concise empathic statements or reflections that
add limited emotional or semantic content without asking, advising, analyzing,
or redirecting. The third includes short utterances that do not serve either
of these functions. The first two categories are included as minimal responses
in the dataset analysis, whereas the third is excluded from the final
statistics.

\begin{figure*}[t]
\centering
\begin{tcolorbox}[
    colback=gray!15,
    colframe=gray!70,
    coltitle=white,
    colbacktitle=gray,
    title=\texttt{Prompt for Contextual Classification},
    fonttitle=\bfseries\sffamily,
    fontupper=\footnotesize,
    boxrule=0.4mm,
    arc=2mm,
    left=2mm,
    right=2mm,
    top=1mm,
    bottom=1mm,
    sharp corners=south,
    enhanced jigsaw
]

{\normalsize\textbf{\# Task}}\\
Classify the interactional function of a candidate counselor utterance in its
specific dialogue context.

\medskip
{\normalsize\textbf{\# Evidence}}\\
Use all three pieces of evidence together:

1. The recent dialogue history before the candidate.\\
2. The candidate counselor utterance itself.\\
3. The next client utterance after the candidate.

Judge the primary interactional function of the candidate in this specific
context.

\medskip
{\normalsize\textbf{\# Labels}}\\

\textbf{1. A backchannel-like response}: A brief acknowledgment that mainly
leaves the conversational floor with the client and contains almost no new
semantic content. Its primary function is to signal listening, receipt, or
permission for the client to continue elaborating the same concern.
Context-dependent examples include ``Mm-hmm.,'' ``I see.,'' ``Yeah.,''
``Okay.,'' ``Right.,'' and ``Sure.''

\smallskip
\textbf{2. A brief empathic or reflective response}: A concise empathic
statement or simple reflection that adds some emotional or semantic content
but does not ask a question, offer advice, analyze the situation, or redirect
the conversation. Examples include ``That sounds really hard.,''
``You felt completely alone.,'' and ``That was frustrating.''

\smallskip
\textbf{3. Another short response}: A short utterance that does not function
as either a backchannel-like response or a brief empathic or reflective
response. This category includes questions, answers, advice, directives,
greetings or closings, logistical expressions, analysis, topic shifts, and
other substantive counselor turns.

\medskip
{\normalsize\textbf{\# Important Exclusions}}\\

1. Do not classify an utterance as a backchannel-like response merely because
it is brief or contains an expression such as ``okay,'' ``yes,'' or
``I see.''\\

2. If the candidate simply answers, accepts, or confirms a question, proposal,
or request in the preceding turn, classify it as another short response.\\

3. If the candidate concerns scheduling, homework, payment, confidentiality,
session opening or closing, thanks, goodbye, or other logistical matters,
classify it as another short response.\\

4. Use the next client utterance as contextual evidence. Continued elaboration
of the same concern may support a backchannel-like response or a brief
empathic or reflective response, but the classification must still be based on
the candidate's own interactional function.\\

5. If the next client utterance instead moves to thanks, closing, scheduling,
agreement, or a new topic, treat this as evidence against a backchannel-like
response.

\medskip
{\normalsize\textbf{\# Input}}\\
\textbf{Recent Dialogue History:}\\
\texttt{\{dialogue\_history\}}

\smallskip
\textbf{Candidate Counselor Utterance:}\\
\texttt{\{candidate\_utterance\}}

\smallskip
\textbf{Next Client Utterance:}\\
\texttt{\{next\_client\_utterance\}}

\medskip
{\normalsize\textbf{\# Output}}\\
Return only compact JSON:

\texttt{\{"label":"backchannel\_like|brief\_empathic\_reflective|other\_short"\}}

\end{tcolorbox}

\caption{
Prompt used by GPT-5.4-mini to contextually classify candidates retained after
rule-based filtering.
}
\label{fig:contextual_classification_prompt}
\end{figure*}

\section{Experimental Details}
\label{app:experimental_details}

\subsection{Models and Inference Prompts}
\label{app:models_prompts}

Figure~\ref{fig:minimal_response_examples} shows concrete examples of the extracted minimal responses.

\subsubsection{Models.} We evaluate the model variants reported in Table~\ref{tab:minimal_response_generation}. The compared systems include general open-source models, an advanced commercial model, counseling-domain fine-tuned models, and human counselor responses. All model-based experiments are conducted in an inference-only setting; we do not update model parameters or perform additional fine-tuning. Unless otherwise specified, we use the original or default inference configurations provided for each model, including decoding and generation parameters.

\begin{itemize}
    \item \textbf{Qwen3-8B} \cite{qwen} We use Qwen3-8B\footnote{\url{https://huggingface.co/Qwen/Qwen3-8B}} as the general open-source baseline for Chinese. This model is used to examine whether a general-purpose instruction-following model can generate minimal responses without counseling-specific fine-tuning.

    \item \textbf{GPT-5.4.} We use GPT-5.4\footnote{\url{https://platform.openai.com/docs/models}} as one of the strongest commercial models. It serves as an upper-bound reference for evaluating whether advanced proprietary LLMs can follow the minimal-response instruction.

    \item \textbf{SoulChat2.0} \cite{xie-etal-2025-psydt} We use SoulChat2.0-Llama-3.1-8B\footnote{\url{https://modelscope.cn/models/YIRONGCHEN/SoulChat2.0-Llama-3.1-8B}}, a counseling-oriented model from the SoulChat2.0 project. This model is included to examine whether models fine-tuned on synthetic counseling data can capture minimal-response behavior.

    \item \textbf{MindChat} We use MindChat-Qwen-7B-v2\footnote{\url{https://huggingface.co/X-D-Lab/MindChat-Qwen-7B-v2}}, a Chinese mental-health dialogue model based on Qwen. It is also used as a representative model fine-tuned for psychological support conversations.

    \item \textbf{KokoroChat-Full} \cite{qi-etal-2025-kokorochat}
We use Llama-3.1-KokoroChat-Full\footnote{\url{https://huggingface.co/UEC-InabaLab/Llama-3.1-KokoroChat-Full}},
which is fine-tuned on Japanese human-collected counseling dialogues.
To avoid overlap between the training and evaluation data, we remove all
source dialogues containing any evaluation example and retrain the model
following the original training setup, as described below.

\item \textbf{Llama 3.1 Swallow} \cite{Fujii:COLM2024, Okazaki:COLM2024}
We use Llama-3.1-Swallow-8B-Instruct-v0.3\footnote{\url{https://huggingface.co/tokyotech-llm/Llama-3.1-Swallow-8B-Instruct-v0.3}}
as the general-purpose open-source baseline for Japanese. This model is
adapted for Japanese instruction following and is used to evaluate
minimal-response generation without counseling-specific fine-tuning.

\item \textbf{PsyDial-Pi4} \cite{qiu-lan-2025-psydial}
We use PsyDial-Pi4\footnote{\url{https://huggingface.co/qiuhuachuan/PsyDial-Pi4}},
which is fine-tuned from Qwen2.5-7B-Instruct on PsyDial-D4.
To avoid overlap between the training and evaluation data, we remove all
source dialogues containing any evaluation example and retrain the model
following the original training setup, as described below.
    
    \item \textbf{Llama-3} \cite{llama3} We use Meta-Llama-3-8B-Instruct\footnote{\url{https://huggingface.co/meta-llama/Meta-Llama-3-8B-Instruct}} as the English open-source baseline. This model is included to test whether a general English instruction-following model can generate minimal responses in contexts extracted from AnnoMI.

\end{itemize}

\subsubsection{Retraining of human-data fine-tuned models.}
To prevent overlap between the training and evaluation data, we retrained the
models fine-tuned on human-collected counseling dialogues after removing the
complete source dialogues containing any evaluation example.

For KokoroChat-Full, the 300 evaluation examples originated from 268 unique
source dialogues. We removed all of these dialogues, together with 108
dialogues from the original KokoroChat test split. The remaining 6,213
dialogues were randomly divided into 5,592 training dialogues and 621
validation dialogues using seed 42. We constructed response-generation
instances from counselor turns, resulting in 192,598 training instances and
21,668 validation instances. We then fine-tuned
Llama-3.1-Swallow-8B-Instruct-v0.3 using QLoRA with 4-bit NF4 quantization.
We used a LoRA rank of 8, an alpha of 16, and a dropout rate of 0.05, and
applied LoRA to the attention and MLP projection modules. The model was trained
for one epoch on eight NVIDIA RTX A6000 GPUs with 48 GB of memory each, using
a maximum sequence length of 4,096, a per-device batch size of 1, gradient
accumulation of 2, and a learning rate of $1\times10^{-3}$.

For PsyDial-Pi4, the 300 evaluation examples originated from 273 unique source
dialogues, all of which were removed before retraining. The remaining 2,109
dialogues were randomly divided into 1,898 training dialogues and 211
validation dialogues using seed 42. We fine-tuned Qwen2.5-7B-Instruct using
full-parameter supervised fine-tuning with FSDP on eight NVIDIA RTX A6000 GPUs
with 48 GB of memory each. The model was trained for two epochs with a maximum
sequence length of 4,096, a per-device batch size of 1, gradient accumulation
of 1, and a learning rate of $1\times10^{-5}$.

\subsubsection{Inference prompts.}
\begin{figure*}[t]
\centering
\includegraphics[width=\textwidth]{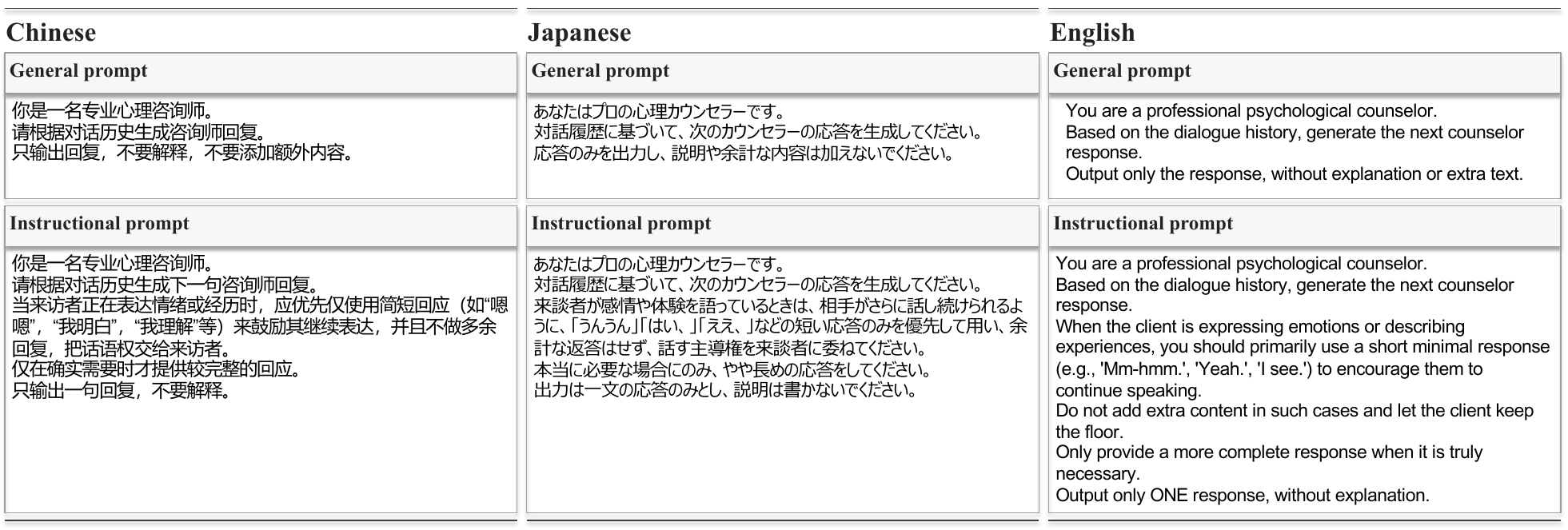}
\caption{Prompt templates used for response generation.}
\label{fig:prompt_templates}
\end{figure*}

We use two prompt settings for the general open-source models and GPT-5.4. The general prompt asks the model to generate the next counselor response based only on the dialogue history. The instructional prompt further encourages the model to use concise minimal responses when the client is still expressing emotions or experiences. Figure~\ref{fig:prompt_templates} shows the prompt templates used for response generation.

\subsection{LLM-based Evaluation Prompts}
\label{app:llm_eval_prompt}

We use GPT-5.4-mini and Gemini-3.1-Flash-Lite as LLM-based evaluators, and
report the average of their scores. Each generated counselor response is first
evaluated from two perspectives. The evaluator determines whether the response
functions as a minimal response in context and assigns an interruption-risk
score from 0 to 5, where a lower score indicates a lower likelihood of
interrupting the client's continued expression. The evaluator considers the
dialogue history, the generated counselor response, and the immediately
following client utterance, and is instructed to return only a JSON object
containing these two fields. Figure~\ref{fig:prompt_eval} shows the prompts used
for this evaluation.

We additionally evaluate general response quality using the response-quality
evaluation prompt presented in Figure~10 of Zhang et al.~
\cite{zhang-etal-2024-cpsycoun}. Responses are scored
along four dimensions: comprehensiveness on a 0--2 scale, professionalism and
authenticity on 0--3 scales, and safety on a 0--1 scale. The overall quality
score is calculated as the average of the four dimension scores.

To support relative rather than isolated evaluation, for each dialogue history
we provide the evaluator with the responses generated by all evaluated models
and ask it to score them together in a single evaluation. This allows the
responses to be compared under the same conversational context and evaluation
criteria. The final score for each response is averaged across GPT-5.4-mini and
Gemini-3.1-Flash-Lite.

\section{Relationship Between LLM Quality Scores and Minimal Responses}
\label{app:quality_correlation}

\begin{figure*}[t]
\centering
\includegraphics[width=\textwidth]{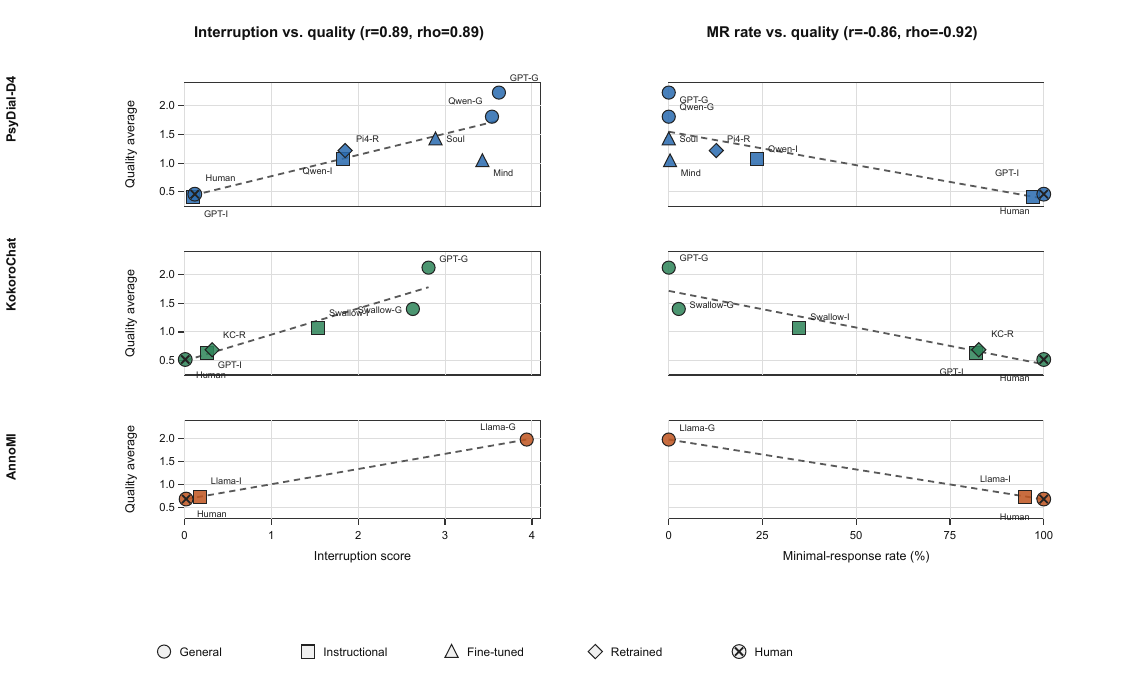}
\caption{
Relationship between LLM-based response-quality scores and interruption scores
or minimal-response rates. Each point represents one dataset--model condition.
Rows correspond to datasets, and columns correspond to the two analyses:
interruption score versus quality score (left) and minimal-response rate versus
quality score (right). Marker shapes indicate model conditions, and dashed
lines show linear trends within each dataset. The correlations shown in the
column titles are computed over all dataset--model conditions.
}
\label{fig:quality_correlation}
\end{figure*}

\begin{figure*}[t]
\centering
\includegraphics[width=\textwidth]{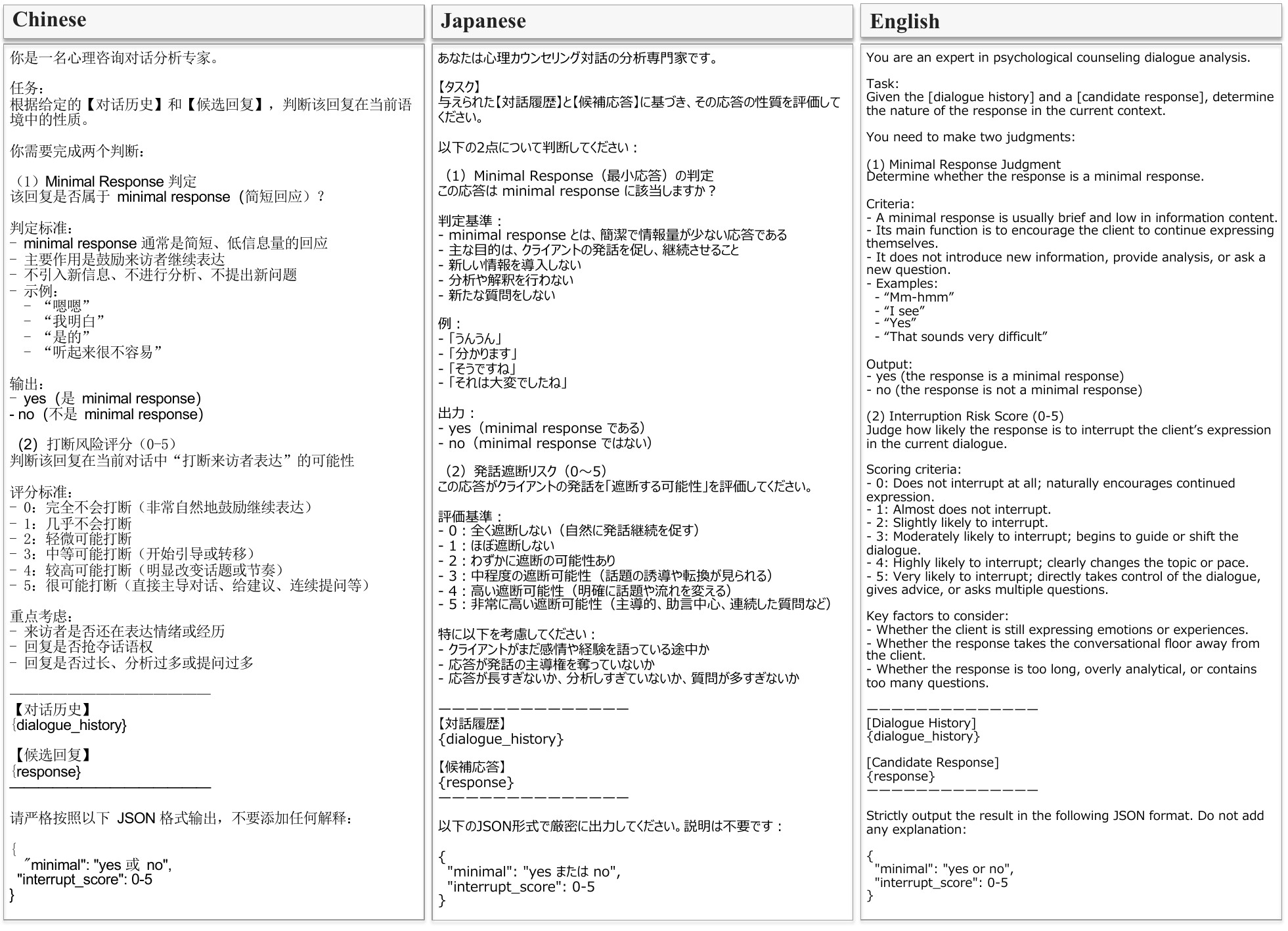}
\caption{
LLM-based evaluation prompts for identifying minimal responses and assessing
interruption risk. The Chinese and Japanese prompts are used in the actual
experiments, while the English version is translated for presentation and is
not used in the experiments.
}
\label{fig:prompt_eval}
\end{figure*}

\begin{figure*}[t]
\centering
\includegraphics[width=\textwidth]{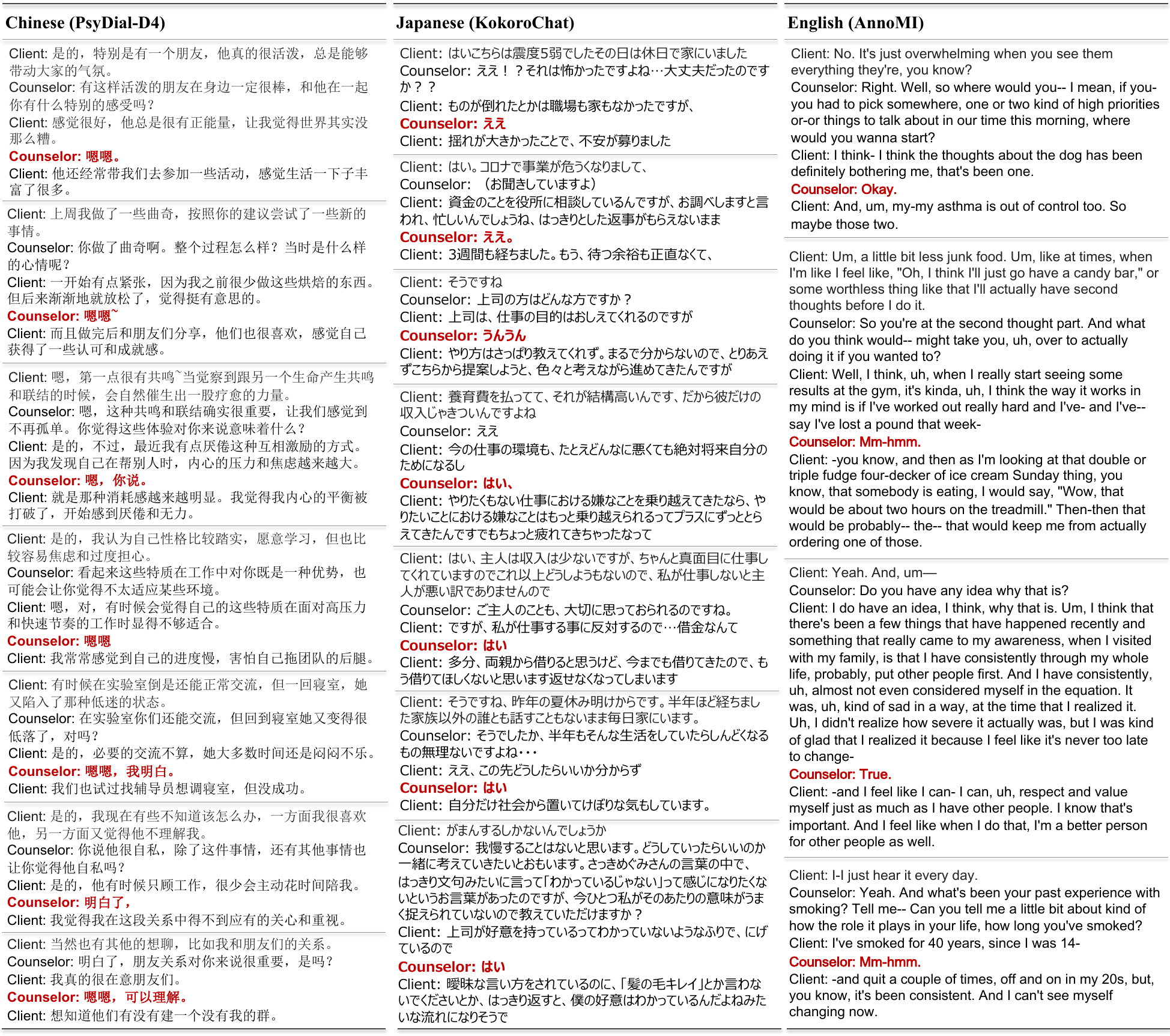}
\caption{Examples of extracted minimal responses in Chinese, Japanese, and English. The bold red utterances are the minimal responses selected by our filtering method. Each example shows a three-turn dialogue history, the counselor's minimal response, and the following client utterance.}
\label{fig:minimal_response_examples}
\end{figure*}

To further examine whether general response-quality evaluation captures the
appropriateness of minimal responses, we analyze the relationship between
LLM-based quality scores, interruption risk, and minimal-response rates.
Each point in Figure~\ref{fig:quality_correlation} represents one
dataset--model condition in the generation experiment. The quality score is
the average score assigned by GPT-5.4-mini and Gemini-3.1-Flash-Lite across
comprehensiveness, professionalism, authenticity, and safety, following the
evaluation dimensions of Zhang et al.~\cite{zhang-etal-2024-cpsycoun}.

As shown in Figure~\ref{fig:quality_correlation}, the LLM-based quality score
is positively correlated with the interruption score (Pearson $r=0.89$; Spearman $\rho=0.89$). In other words, conditions
whose responses are judged more likely to interrupt the client's continued
expression tend to receive higher general response-quality ratings. Conversely,
quality scores are negatively correlated with minimal-response rates
(Pearson $r=-0.86$; Spearman $\rho=-0.92$), suggesting that conditions that
produce minimal responses more frequently tend to receive lower ratings under
this general quality rubric.

These results are consistent with our concern that general-purpose
LLM-as-a-judge evaluation may undervalue minimal responses. Minimal responses
are often brief and contain little explicit informational content; therefore,
they may be penalized by rubrics that emphasize comprehensiveness or
professional elaboration, even when they are contextually appropriate for
preserving the client's narrative space.

\end{document}